\documentclass[nohyperref]{article}

\usepackage{microtype}
\usepackage{graphicx}
\usepackage{subfig}
\usepackage{booktabs} 
\graphicspath{{Figures/}}

\usepackage{amsmath}
\usepackage{amssymb}
\usepackage{mathtools}
\usepackage{amsthm}

\usepackage{caption}
\usepackage{multirow}
\usepackage{setspace}
\usepackage{scalerel}
\usepackage{color}
\usepackage{makecell}
\usepackage{placeins}
\usepackage[bottom]{footmisc}
\usepackage{algorithm,algorithmic}
\usepackage{bm}
\usepackage{float}

\newlength\myindent
\theoremstyle{plain}

\theoremstyle{definition}

\theoremstyle{remark}

\usepackage[disable,textsize=tiny]{todonotes}

\definecolor{citation_color}{RGB}{62, 73, 131}
\usepackage[colorlinks=true,citecolor={citation_color}]{hyperref}%

\usepackage{hyperref}
\usepackage{url}

\usepackage[accepted]{icml2022}

\icmltitlerunning{Stochastic LWTA Networks for Model-Agnostic Meta-Learning}

\begin{document}

\twocolumn[
\icmltitle{Stochastic Deep Networks with Linear Competing Units \\
           for Model-Agnostic Meta-Learning}




\begin{icmlauthorlist}
\icmlauthor{Konstantinos Kalais}{to}
\icmlauthor{Sotirios Chatzis}{to}
\end{icmlauthorlist}

\icmlaffiliation{to}{Dept. of Electrical Eng., Computer Eng., and Informatics, Cyprus University of Technology, Limassol, Cyprus}

\icmlcorrespondingauthor{Konstantinos Kalais}{ki.kalais@cut.ac.cy}
\icmlcorrespondingauthor{Sotirios Chatzis}{sotirios.chatzis@cut.ac.cy}

\icmlkeywords{Meta-Learning, Stochastic Representations, Local Winner-Takes-All}

\vskip 0.3in
]



\printAffiliationsAndNotice{}  

\begin{abstract}
This work addresses meta-learning (ML) by considering deep networks with stochastic local winner-takes-all (LWTA) activations. This type of network units results in sparse representations from each model layer, as the units are organized into blocks where only one unit generates a non-zero output. The main operating principle of the introduced units rely on stochastic principles, as the network performs posterior sampling over competing units to select the winner. Therefore, the proposed networks are explicitly designed to extract input data representations of sparse stochastic nature, as opposed to the currently standard deterministic representation paradigm. Our approach produces state-of-the-art predictive accuracy on few-shot image classification and regression experiments, as well as reduced predictive error on an active learning setting; these improvements come with an immensely reduced computational cost. Code is available at: \url{https://github.com/Kkalais/StochLWTA-ML}
\end{abstract}

\section{Introduction}
\label{intro}
When we train machine learning models on problems with limited amounts of training data, we cannot usually get good predictive performance \citep{Sculley, Lai}. This comes in contrast to the human ability to quickly derive information from a range of different tasks, and then adapt to a new task with limited available new examples \citep{Castro, Kuhl}. In essence, this capability of the mind of learning how to learn \citep{Black} has inspired researchers to investigate the concept of Meta-Learning (ML) \citep{Lake, Matthias}. 

There is a large variety of deep learning methods for ML \citep{Marcin, Finn2, Sungyong}. Specifically, \citet{Finn} presented the \textit{Model-Agnostic Meta-Learning} (MAML) algorithm that enables tuning the parameters of a trained network to quickly learn a new task with only a few gradient updates. To get away with the entailed second-order computations, that are expensive, several researchers proposed appropriate first-order approximations for MAML; such works are the \textit{First-Order MAML} (FOMAML) of \citet{Finn} and the \textit{Reptile} algorithm of \citet{Nichol}. Recently, several researchers have also considered Bayesian inference-driven methods for deep learning ML, extending upon older works built for conventional machine learning approaches \citep{Grant_Finn, Yoon, Finn_Xu, Ravi_Beatson, Patacchiola, Zou_Lu, Chen}. 

This work proposes a different regard toward improving generalization capacity for deep networks in the context of ML. Specifically, our proposed approach relies on the following main concepts:
\begin{itemize}
\item \textbf{The concept of sparse learned representations.} For the first time in the literature of deep network-driven ML, we employ a mechanism that inherently learns to extract sparse data representations. This consists in replacing standard unit nonlinearities (e.g., ReLU) with a unit competition mechanism. Specifically, (linear) units are organized into blocks. Presented with some input, the units within a block engage in a competition process with only one winner. The outputs of all units except for the winner are zeroed out; the output of the winner retains its computed value (local winner takes-all, LWTA, architecture).
    
\item \textbf{The concept of stochastic representations.} We establish a stochastic formulation for the previously described competition process. Specifically, we postulate that, within a block of competing units, winner is selected via sampling from an appropriate Categorical posterior. The corresponding winning probability of each unit is proportional to its linear computation (thus depending on the layer input). Via this competition process, we yield stochastic representations from network layers, that is representations that may change each time we present to the network layer exactly the same input.
    
\end{itemize}

Based on the results from existing approaches, we posit that the proposed treatment of the ML problem, which combines learned representation sparsity and stochasticity, will be extremely beneficial to the deep learning community. We dub our approach Stochastic LWTA for ML (StochLWTA-ML).

We perform a variational Bayes treatment of the proposed model. We opt for a full Bayesian treatment, by also handling network weights as latent variables. That is, we elect to impose an appropriate prior over the network weights and fit approximate (variational) posteriors. As we evaluate our approach on image classification, sinusoidal regression and active learning problems, we show that StochLWTA-ML offers a variety of advantages over the current state-of-the-art methods, namely: (i) incurring reduced predictive error rate compared to the currently state-of-the-art methods in the field; (ii) obtaining this performance with networks that comprise \emph{an immensely reduced} number of trainable parameters, and therefore give rise to better computational efficiency and imposed memory footprint.

The remainder of this paper is organized as follows: In Section \ref{prelims}, we briefly review related work. Section \ref{approach} introduces our approach and provides the related training and prediction algorithms. In Section \ref{experiments}, we perform a thorough experimental evaluation of StochLWTA-ML, and compare our findings to the current state-of-the-art. In the final Section \ref{conclusion}, we end up with the conclusions of our work, and suggest lines of further research.

\section{Related Work}
\label{prelims}

\subsection{LWTA layers in Deep Learning}

LWTA layers are not new in the field of deep learning; see, e.g., \citet{Rupesh}. Although not much work has been pursued along these lines, the recent works of \citet{Panousis_Chatzis, Panousis_Chatzis_Adv} and \citet{Voskou_Chatzis} have spurred some fresh interest in the field. These works have presented alternative implementations of the basic concepts of LWTA units in the context of diverse deep network architectures. Specifically, \citet{Panousis_Chatzis} propose a stochastic LWTA formulation which is founded upon the Indian Buffet process (IBP) prior, borrowed from nonparametric statistics; they use this architecture to effect data-driven network compression. \citet{Panousis_Chatzis_Adv} exploit the same technique to train adversarially-robust deep networks. On the other hand, \citet{Voskou_Chatzis} consider a different incarnation of stochastic LWTA architectures, which relies on sampling the winner from a Categorical posterior, driven from the layer input. This layer architecture is used to replace dense ReLU layers in Transformer networks; it is then shown to yield important benefits in a Sign-Language Translation benchmark.

This paper is different from the previous works in various ways: (i) stochasticity does not stem from utilization of the IBP; we rather adopt an approach similar to \citet{Voskou_Chatzis}; (ii) we do not use the proposed architecture as a replacement for a specific type of layer in a greater network architecture (Transformer) that remains largely unchanged; instead, we build completely new networks using these layers; (iii) we perform a full Bayesian treatment, by treating network weights as random variables; we do not employ this construction as a means of compressing the weights at prediction time, contrary to \citet{Panousis_Chatzis},  \citet{Voskou_Chatzis}; instead, we perform weight sampling at prediction time as a means of improving accuracy; and (iv) for the first time, we examine how these principles perform in the context of deep network-driven ML. Note that, apart from LWTA architectures, other data-driven sparsity models have also been proposed recently, e.g. \citet{Lee_Kim}, \citet{Kessler}. However, none of these have been developed or evaluated in the context of an ML setting.

\subsection{Model-Agnostic Meta-Learning}
\label{maml}
\citet{Finn} suggested a \textit{model-agnostic} algorithm for ML, that can be applied to any model trained via gradient descent. The introduced MAML algorithm initializes model parameters in a way that can be quickly adapted to several types of new tasks. 

Let us consider a model with parameters $\bm{\theta}$ and a parametric form $f_{\bm{\theta}}$. When the model is adapting to an unseen task $T_i$, MAML runs few steps of gradient descent that yields a task-specific parameter set, $\bm{\theta}'_i = \bm{\theta} - \alpha \nabla_{\bm{\theta}}L_{T_i}(f_{\bm{\theta}})$; $\alpha$ is the step size hyperparameter and $L_{T_i}$ denotes the loss on the task $T_i$. Subsequently, training proceeds to optimize the function $f_{\bm{\theta}'_i}$ with respect to the model parameters $\bm{\theta}$. Assuming that the batch of tasks has size $M$, we can define the targeted \textit{meta-objective} as:
\begin{equation} \label{eq:1}
L_{meta}(\bm{\theta}) = \sum_{i=1}^M L_{T_i}(f_{\bm{\theta}'_i}) = \sum_{i=1}^M L_{T_i}(f_{\bm{\theta}- \alpha \nabla_{\bm{\theta}}L_{T_i}(f_{\bm{\theta}})}).
\end{equation}
Optimization of this meta-objective over $\bm{\theta}$ yields the \textit{outer-loop} update:
\begin{equation} \label{eq:2}
\bm{\theta} \leftarrow \bm{\theta} - \beta \nabla_{\bm{\theta}}\sum_{i=1}^M L_{T_i}(f_{\bm{\theta}'_i})
\end{equation}
where $\beta$ stands for the \textit{outer-loop} learning rate.

Finally, as \textit{MAML} involves expensive computations stemming from the second-order updates of Eqs. (\ref{eq:1}) and (\ref{eq:2}), \citet{Finn} and \citet{Nichol} have developed more efficient first-order approximations.

\section{Proposed Approach}
\label{approach}

\subsection{Architecture}
\label{model_def}
Let us denote as $\bm{x} \in \mathbb{R}^I$ an input vector presented to a dense ReLU layer of a conventional deep neural network, with corresponding weights matrix $\bm{W} \in \mathbb{R}^{I \times O}$. The output of the layer is the vector $\bm{y} \in \mathbb{R}^O$  and is fed to the subsequent layer. In our approach, a ReLU unit is replaced by $J$ competing linear units, organized in one block; in the following, we denote with $R$ the number of blocks in a layer. The input $\boldsymbol{x}$ is now presented to each block through weights that are organized into a three-dimensional matrix $\bm{W} \in \mathbb{R}^{I \times R \times J}$. Then, the $j$-th competing unit within $r$-th block computes the sum $\sum_{i=1}^{I} (w_{i,r,j}) \cdot x_i$. Competition means that, of the $J$ units in the block, one unit (the "winner") will present its linear computation to the next layer; the rest will present zero values. Traditionally in the literature, the winner unit is selected to be the unit with greatest linear computation. Recently, stochastic competition principles have been considered, e.g. \citet{Panousis_Chatzis, Panousis_Chatzis_Adv}, \citet{Voskou_Chatzis}. 

Let us denote as $\bm{y} \in \mathbb{R}^{R \cdot J}$ the output of an LWTA layer; this is composed of $R$ subvectors $\bm{y}_r \in \mathbb{R}^J$ and is sparse, since all units except for one, in each block, yield zero values. Let us introduce the discrete latent indicator vector $\bm{\xi} \in \mathrm{one\_hot}(J)^R$ to denote the winner units in the $R$ blocks that constitute a considered stochastic LWTA layer. This vector comprises $R$ component subvectors, where each component entails one non-zero value at the index position that corresponds to the winner unit of the respective LWTA block. On this basis, the output $\bm{y}$ of the stochastic LWTA layer's $(r \cdot j)$-th component $\bm{y}_{r,j}$ is defined as:
\begin{equation} \label{eq:3}
\bm{y}_{r,j} = \bm{\xi}_{r,j} \sum_{i=1}^{I} (w_{i,r,j}) \cdot x_i \in \mathbb{R}
\end{equation}
where we denote as $\bm{\xi}_{r,j}$ the $j$-th component of $\bm{\xi}_r$, and 
$\bm{\xi}_r \in \mathrm{one\_hot}(J)$ holds the $r$-th subvector of $\bm{\xi}$.

\begin{figure*}[t!]
  \centering
  \includegraphics[width=0.9\linewidth]{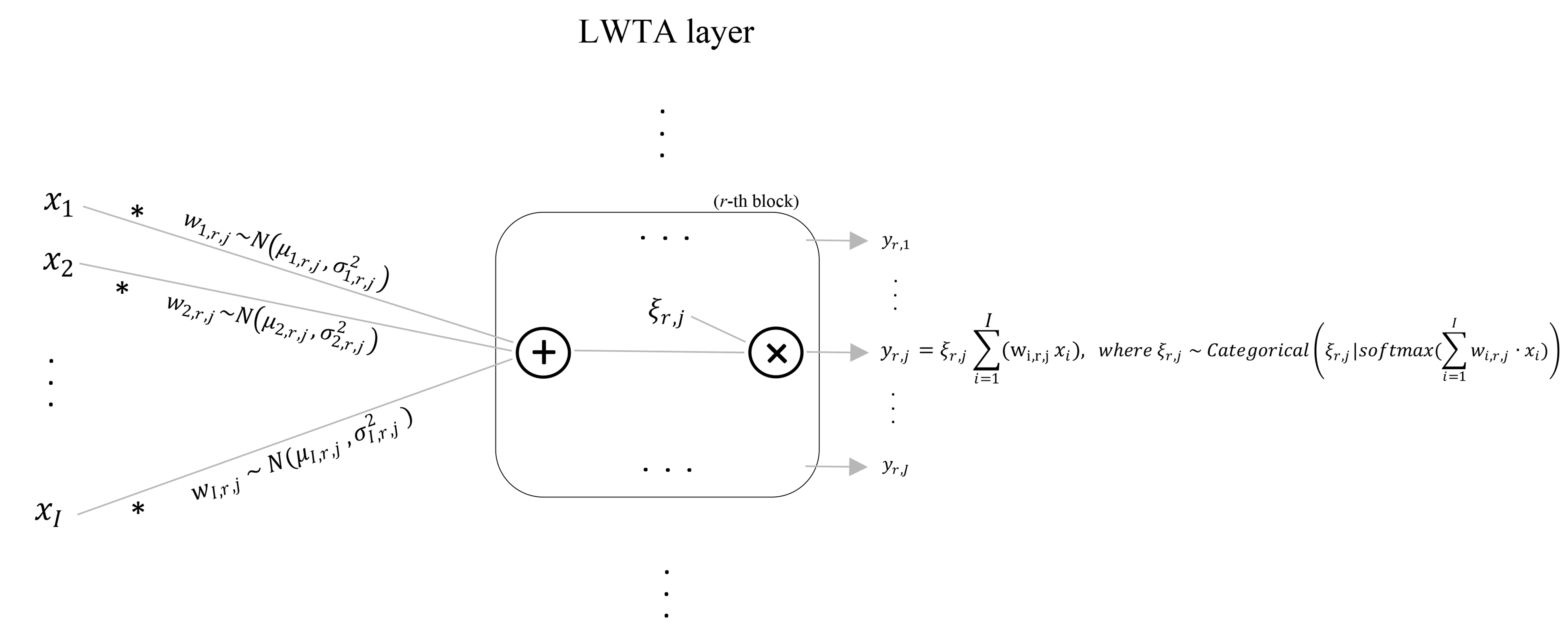} 
  \caption{A zoomed-in graphical illustration of the $r$-th block of a stochastic LWTA layer. Input $\boldsymbol{x} = [x_1, x_2, \ldots, x_{I}]$ is presented to each unit in the block. The $j$-th component $\bm{y}_{r,j}$ of the block's output is obtained after computing latent variable $\bm{\xi}_{r,j}$.}
  \label{fig:model}
  \vskip -0.1in
\end{figure*}

In Eq. (\ref{eq:3}), we postulate that the latent winner indicator variables are drawn from a Categorical distribution which is proportional to the intermediate linear computation that each unit performs. Therefore, the stronger the linearity the higher the chance of the unit winning the stochastic competition within its block. In detail, we postulate that, a posteriori, the winner distributions yield: 
\begin{equation} \label{eq:4}
q(\bm{\xi}_r) = \mathrm{Categorical}\left(\bm{\xi}_r \scaleobj{2}{\mid} \mathrm{softmax}(\sum_{i=1}^{I}[w_{i,r,j}]_{j=1}^J \cdot x_i) \right)
\end{equation}
where $[w_{i,r,j}]_{j=1}^J$ denotes the vector concatenation of the set $\{w_{i,r,j}\}_{j=1}^J$. A graphical illustration of the proposed stochastic architecture is provided in Fig. \ref{fig:model}.

As a network composed of such (StochLWTA) layers entails latent variables $\boldsymbol{\xi}$, we need to perform a Bayesian network treatment to perform effective parameter training. We opt for a (approximate) stochastic gradient variational Bayes treatment \citep{Kingma}, for scalability purposes. This means that the used objective function takes the form of an evidence lower-bound (ELBO) objective, as we describe next. In our work, we take one step further: we also elect to infer a posterior density over the network weights $\boldsymbol{W}$, as opposed to obtaining point-estimates. This results in a second source of stochasticity for our approach, which may further increase its generalization capacity under uncertain conditions arising from limited training data availability. Note that the use of these posteriors is totally different from \citet{Panousis_Chatzis} and \citet{Voskou_Chatzis}: therein, posterior variance is used for compressing posterior mean bit-precision; then, predictions are performed using only the compressed posterior mean. Instead, in our work we sample multiple times from the trained weight posterior and perform model averaging (in a Bayesian sense), as a means of increasing generalization capacity.

We postulate:
\begin{equation} \label{eq:5}
q(\mathrm{vec}(\bm{W})) = N(\mathrm{vec}(\bm{W})|\bm{\mu}, \mathrm{diag}(\bm{\sigma}^2))
\end{equation}
where $\bm{\psi} \triangleq \{\bm{\mu}, \bm{\sigma}^2\}$ are the means and variances of the Gaussian weight posteriors, respectively.

\subsection{A Model-Agnostic ML Algorithm}
\label{training}
Let us define an ML problem where we are given a training dataset $D$ of tasks, $T$, that are governed by a distribution $P(T)$. We consider an $N$-way, $K$-shot learning setting, where each task contains $K$ labelled examples for each of $N$ available classes. 

Our approach entails parameter sets $\boldsymbol{\theta}$ which coincide with the hyperparameter sets $\bm{\psi} = \{\bm{\mu}, \bm{\sigma}^2\}$ of the weights $\boldsymbol{W} \in \mathbb{R}^{I \times R \times J}$; these sets $\bm{\psi} = \{\bm{\mu}, \bm{\sigma}^2\}$ are the target of our MAML-type training algorithm. 

Therefore, to perform model training, we have to first initialize the trainable parameters $\bm{\mu}$ and $\bm{\sigma}^2$ across layers. To this end, one can appropriately exploit popular initialization schemes, such as Glorot Uniform \citep{glorot}. Then, our approach entails an inner-outer loop scheme, in a vein similar to existing approaches, but with some crucial differences: 

\begin{enumerate}

\item The stochastic nature of the postulated weights, $\boldsymbol{W}$, results in the updates taking place over the posterior means, $\boldsymbol{\mu}$ and variances, $\boldsymbol{\sigma}^2$.

\item The stochastic nature of both the inferred representations and the network weights themselves implies that proper training must rely on optimization of the ELBO function of the network. Let us consider an inner-loop dealing with task $T_i \sim P(T)$, with data $D_i=(X_i,Y_i) \subset D$. Let $\mathrm{CE}(Y_i, f_{\bm{\psi}}(X_i; \hat{\boldsymbol{\xi}}, \hat{\boldsymbol{W}}))$ be the categorical cross-entropy between the data labels $Y_i$ and the class probabilities $f_{\bm{\psi}}(X_i;\hat{\boldsymbol{\xi}},\hat{\boldsymbol{W}})$ generated by the Softmax layer. Then, we yield: 
\begin{equation} \label{eq:6}
\begin{split}
L_{T_i}(\bm{\psi}) = &-\mathrm{CE}(Y_i, f_{\bm{\psi}}(X_i; \hat{\boldsymbol{\xi}}, \hat{\boldsymbol{W}})) \\
& -KL[\,q(\bm{\xi})\,||\,p(\bm{\xi})\,] - KL[\,q(\bm{W})\,||\,p(\bm{W})\,]
\end{split}
\end{equation}
for task $T_i$ with dataset $D_i$ which is dealt with on the $i$-th training iteration. 
\end{enumerate}

Here, for simplicity and without harming generality, we consider that the weights prior $p(\mathrm{vec}(\bm{W}))$ is a Gaussian distribution $N(\bm{0},\bm{I})$ and the latent variables prior $p(\bm{\xi})$ is a symmetric Categorical distribution $\mathrm{Categorical}(1/J)$. In addition, in our notation we stress that the output $f_{\bm{\psi}}(X_i;\hat{\boldsymbol{\xi}},\hat{\boldsymbol{W}})$ depends on the winner selection process, which is stochastic, and the outcomes of sampling the network weights. Specifically, in our work we perform Monte-Carlo (MC) sampling using one reparameterized sample of the corresponding latent variables. 

Let $\bm{\bar{\xi}}$ be the unnormalized probabilities of the Categorical distribution $q(\bm{\xi})$ (Eq. (\ref{eq:4})). The sampled instances of $\bm{\xi}$, $\hat{\bm{\xi}}_{r,j}$, are expressed as \citep{Maddison}:
\begin{equation} \label{eq:7}
\begin{split}
&\hat{\bm{\xi}}_{r,j} = \mathrm{Softmax}((\log\bar{\bm{\xi}}_{r,j} + g_{r,j})/\tau) , \; \\
& \forall r=1, \dots, R, \,j=1, \dots, J
\end{split}
\end{equation}
where $g_{r,j} = -\log(\,-\log U_{r,j}), \, U_{r,j} \sim \mathrm{Uniform}(0,1)$, and $\tau \in (0,\infty)$ is a temperature factor that controls how closely the Categorical distribution is approximated by this continuous relaxation.

Similarly, the Gaussian weights yield:
$\hat{w}_{t,r,j} = \mu_{t,r,j} + \sigma_{t,r,j} \hat{\epsilon}, \,$ and $\hat{\epsilon} \sim N(0,1)$.\par

On this basis, the KL divergences in Eq. (\ref{eq:6}) become:
\begin{equation} \label{eq:8}
\begin{split}
& KL[\,q(\bm{\xi}_{r,j}\,||\,p(\bm{\xi}_{r,j})\,] = \mathbb{E}_{q(\bm{\xi}_{r,j})}[\log q(\bm{\xi}_{r,j}) - \log p(\bm{\xi}_{r,j})]\\ & \approx \log q(\hat{\bm{\xi}}_{r,j}) - \log p(\hat{\bm{\xi}}_{r,j}), \; \forall r, \, j
\end{split}
\end{equation}
and
\begin{equation} \label{eq:9}
\begin{split}
& KL[\,q(w_{t,r,j})\,||\,p(w_{t,r,j})\,] = \mathbb{E}_{q(w_{t,r,j})}[\log q(w_{t,r,j}) - \\
& \log p(w_{t,r,j})] \approx \log q(\hat{w}_{t,r,j}) - \log p(\hat{w}_{t,r,j}), \; \\
& \forall t=1, \dots, I, \,r , \, j
\end{split}
\end{equation}

Hence, the ELBO becomes:
\begin{equation} \label{eq:10}
\small
\begin{split}
& L_{T_i}(\bm{\psi}) = -\mathrm{CE}(Y_i, f_{\bm{\psi}}(X_i; \hat{\boldsymbol{\xi}},\hat{\boldsymbol{W}})) - \sum_{r,j} (\log q(\hat{\bm{\xi}}_{r,j}) - \\
& \log p(\hat{\bm{\xi}}_{r,j}) )  - \sum_{t,r,j} \left(\log q(\hat{w}_{t,r,j}) - \log p(\hat{w}_{t,r,j}) \right)
\end{split}
\end{equation}

Therefore, we establish a MAML-type algorithm, where: (i) the used networks comprise blocks of stochastic LWTA units visually depicted in Fig. \ref{fig:model}; (ii) the trainable parameters are the means $\bm{\mu}$ and variances $\bm{\sigma}^2$ of the synaptic weights; and (iii) the objective function of the inner-loop process is given in Eq. (\ref{eq:10}).

We summarize our training algorithm in Alg. \ref{alg:alg1}.

\begin{algorithm}[h!]
 \caption{Model training with StochLWTA-ML}
 \label{alg:alg1}
\begin{algorithmic}
  \STATE {\bfseries Require:} $P(T)$: distribution over tasks
  \STATE Initialize $\bm{\psi}:=\{\bm{\mu},\bm{\sigma}^2\}$ 
  \STATE Define outer-step size $\beta$ and inner learning rate $\alpha$ 
  \FOR{i = 1,2, $\ldots$}
  \STATE \textbf{Inner training loop}:
  \STATE \hspace{0.1in} Sample task $T_i \sim P(T)$
  \STATE \hspace{0.1in} Compute $L_{T_i}(\bm{\psi})$ using Eq. (\ref{eq:10})
  \STATE \hspace{0.1in} Compute adapted parameters with SGD: $\bm{\psi}'_i = \bm{\psi} -$
  \STATE \hspace{0.1in} $\alpha \nabla_{\bm{\psi}}L_{T_i}(f_{\bm{\psi}})$
  \STATE \textbf{Outer training loop}:
  \STATE \hspace{0.1in} Derive $\bm{\psi} \leftarrow \bm{\psi} + \beta(\bm{\psi}'_i - \bm{\psi})$
  \ENDFOR
\end{algorithmic}
\end{algorithm}

\subsection{Prediction Algorithm}
\label{infer}
We start our prediction algorithm (Alg. \ref{alg:alg2}) by sampling a task $T' \sim P(T)$, with data $D'=(X',Y') \subset D$. Then, we draw a set of $B$ samples of the Gaussian connection weights from the trained posteriors $\mathcal{N}(\boldsymbol{\mu}, \boldsymbol{\sigma}^2)$, and select the winning units in each block of the network by similarly sampling from the posteriors $q(\bm{\xi})$. This results in a set of $B$ output logits of the network, which we average to obtain the final predictive outcome:
\begin{equation} \label{eq:11}
f_{\bm{\psi}}(X';\tilde{\boldsymbol{\xi}},\tilde{\boldsymbol{W}}) \approx \frac{1}{B} \sum_{s=1}^B f_{\bm{\psi}}(X';\tilde{\boldsymbol{\xi}_s},\tilde{\boldsymbol{W}_s}) 
\end{equation}
where $\tilde{\boldsymbol{\xi}_s}$ and $\tilde{\boldsymbol{W}_s}$ are sampled directly from the posteriors $q(\bm{\xi})$ and $q(\bm{W})$, respectively.

This concludes the formulation of the proposed model-agnostic ML approach.
\begin{algorithm}[h!]
  \caption{Prediction with StochLWTA-ML}
  \label{alg:alg2}
\begin{algorithmic}
    \STATE {\bfseries Require:} Learned parameters $\bm{\psi}=\{\bm{\mu}, \bm{\sigma}^2\}$
    \STATE Sample task $T' \sim P(T)$
    \FOR{$s=1$ {\bfseries to} $B$}
        \STATE Sample $\tilde{\boldsymbol{W}_s} \sim q(\bm{W})$ defined in Eq. (\ref{eq:5})
        \STATE Sample $\tilde{\boldsymbol{\xi}_s} \sim q(\bm{\xi})$ defined in Eq. (\ref{eq:4}), for $\bm{W}=\tilde{\bm{W}}$ and $(x_i=x_i') \in X'$
        \STATE Compute output logits $f_{\bm{\psi}}(X';\tilde{\boldsymbol{\xi}_s},\tilde{\boldsymbol{W}_s})$, given the sampled values $\tilde{\boldsymbol{\xi}_s}$ and $\tilde{\boldsymbol{W}_s}$
    \ENDFOR
    \STATE Use Eq. (\ref{eq:11}) to average over the resulting set of $B$ logits and derive the final prediction.
\end{algorithmic}
\end{algorithm}

\section{Experiments}
\label{experiments}

We evaluate our proposed model in various few-shot learning tasks: image classification, sinusoidal regression and active learning. After thorough exploration on the number of LWTA layers as well as the number of blocks for each layer and the competing units per block, we end up with using networks comprising 2 layers with 16 blocks and 2 competing units per block on the former layer, and 8 blocks with 2 units per block on the latter. The last network layer is a Softmax. Weight mean initialization, as well as point-estimate initialization for our competitors, is performed via Glorot Uniform. Weight log-variance initialization is performed via Glorot Normal, by sampling from $N(0.0005, 0.01)$. The Gumbel-Softmax relaxation temperature is set to $\tau = 0.67$. 

In the inner-loop updates, we use the Stochastic Gradient Descent (SGD) \citep{Robbins} optimizer with a learning rate of 0.003. For the outer-loop, we use SGD with a linear annealed outer step size to 0, and an initial value of 0.25. Additionally, all the experiments were ran with task batch size of 50 for both training and testing mode. Prediction is carried out averaging over $B=4$ output logits. The code was implemented in Tensorflow \citep{Abadi}.

\subsection{Classification}
\label{classification}
We first evaluate StochLWTA-ML on popular few-shot image classification datasets, and compare its performance to state-of-the-art prior results. In Table \ref{table1}, we show how StochLWTA-ML performs on Omniglot 20-way \citep{Lake}, Mini-Imagenet 5-way \citep{Vinyals} and CIFAR-100 5-way \citep{Krizhevsky_cifar} few-shot settings. We compare our findings to state-of-the-art ML algorithms such as LLAMA and PLATIPUS as reported in \citet{Gordon_Bronskill}, Amortized Bayesian Meta-Learning (ABML), MAML, FOMAML, Reptile and others. Using the original architectures with the same hyperparameters and data preprocessing as in \citet{Finn}, we have also locally reproduced ABML, BMAML (with 5 particles), PLATIPUS, MAML, FOMAML and Reptile (dubbed "local" in Table \ref{table1}). 

For the local replicates, the results in Table \ref{table1} constitute average performance statistics and corresponding standard deviations (std's) over three runs, using different random seeds. For completeness sake, we also compare our findings to other state-of-the-art ML models as reported in \citet{Finn}, including Matching Nets and LSTM Meta-Learner. As we observe, our method outperforms the existing state-of-the-art in both the 1-shot and 5-shot settings. Besides, the reported std statistics on the locally reproduced experiments show consistency across all methods; thus, our stochastic approach is as resilient to seed initialization as the existing approaches.

\begin{table*}[h!]
\centering
\captionsetup{justification=centering}
\caption{N-way K-shot $(\%)$ classification accuracies on Omniglot, Mini-Imagenet and CIFAR-100}
\label{table1}
\vskip 0.1in
\small
\begin{tabular}{lcccccc}  
\toprule
&  \multicolumn{2}{c}{\textbf{\makecell{Omniglot 20-way}}} &  \multicolumn{2}{c}{\textbf{\makecell{Mini-Imagenet 5-way}}} & \multicolumn{2}{c}{\textbf{\makecell{CIFAR-100 5-way}}}\\
\midrule
\textbf{Algorithm}   &  1-shot     &  5-shot    &  1-shot        &  5-shot   & 1-shot & 5-shot \\
\midrule
Matching Nets  \citep{Santoro}  & 93.80  & 98.50 & 43.56  & 55.31 & - & -\\
LSTM Meta-Learner \citep{Ravi}   &  -  & - & 43.44  & 60.60   & - & -\\
MAML                   &  95.80  &  98.90 & 48.70  & 63.11  & - & -\\
FOMAML                   &  -  &  - & 48.07  & 63.15  & - & -\\
Reptile                 &  88.14  &  96.65 & 47.07  & 62.74   & - & -\\
PredCP \citep{Nalisnick}        &  -  &  - & 49.30  & 61.90  & - & -\\
Neural Statistician \citep{Harrison} &  93.20  &  98.10 & -  & -   & - & -\\
mAP-SSVM  \citep{Triantafillou} &  95.20 & 98.60 & 50.32  & 63.94   & - & -\\
LLAMA     \citep{Grant_Finn}       &  -  & - & 49.40  & -   & - & -\\
PLATIPUS    \citep{Finn_Xu}      &  -  & - & 50.13  & -  & - & -\\
GEM-BML+  \citep{Zou_Lu}        & 96.24  & 98.94 & 50.03  & -  & - & -\\
DKT \citep{Patacchiola}         &  -  & - & 49.73  & 64.00  & - & -\\
ABML \citep{Ravi_Beatson} &  -  & - & 45.00  & -   & 49.50 & -\\
BMAML (with 5 particles) \citep{Yoon}   &  -  & - & 53.80 & -   &  & -\\
\midrule
ABML (local)          &  90.21 \tiny ± 0.34 & 93.39 \tiny ± 0.09 & 44.23 \tiny ± 0.81 & 52.12 \tiny ± 1.01  & 49.23 \tiny ± 0.23 & 53.60 \tiny ± 0.39\\
BMAML (local)          &  96.92 \tiny ± 0.58 & 98.11 \tiny ± 2.03 & 53.10 \tiny ± 1.05 & 64.80 \tiny ± 0.93 & 52.60 \tiny ± 1.40 & 65.80 \tiny ± 0.04\\
PLATIPUS (local)       &  94.35 \tiny ± 0.87 & 98.30 \tiny ± 0.44 & 49.97 \tiny ± 0.97 & 63.13 \tiny ± 1.18 & 51.14 \tiny ± 0.48 & 63.61 \tiny ± 2.16\\
MAML (local)          &  95.48 \tiny ± 0.81 & 98.61 \tiny ± 0.49 & 48.60 \tiny ± 1.23 & 63.01 \tiny ± 1.28 & 50.67 \tiny ± 0.93 & 62.89 \tiny ± 0.77\\
FOMAML (local)        &  94.92 \tiny ± 0.71 &  98.12 \tiny ± 0.94 & 47.93 \tiny ± 0.67 & 63.10 \tiny ± 0.58 & 49.13 \tiny ± 0.61 & 63.80 \tiny ± 0.83 \\
Reptile (local)         &  87.98 \tiny ± 1.18 &  96.36 \tiny ± 1.54 & 46.97 \tiny ± 0.95 & 62.53 \tiny ± 0.61 & 48.19 \tiny ± 0.74 & 63.45 \tiny ± 1.63 \\
\midrule
\textbf{StochLWTA-ML}              &  \textbf{97.79 \tiny ± 0.48}  &  \textbf{98.97 \tiny ± 0.61} & \textbf{54.11 \tiny ± 0.82}  & \textbf{66.70 \tiny ± 0.41}   & \textbf{54.60 \tiny ± 0.39} & \textbf{66.73 \tiny ± 0.06}\\
\bottomrule
\end{tabular}
\vskip -0.1in
\end{table*}

In the following, we perform a diverse set of ablations that shed light to the characteristics and capabilities of our approach. Few more ablations are deferred to Appendix D.

\subsubsection{Does stochastic competition contribute to classification accuracy?}
\label{variants}

To check whether the accuracy improvements stem from the LWTA-induced sparsity or the proposed stochastic competition concept, we evaluate both our approach as well as MAML, FOMAML, ABML, BMAML and PLATIPUS, considering both "deterministic LWTA" and "stochastic LWTA" setups; deterministic LWTA networks have been adopted from \citet{Rupesh}. As we see in Table \ref{table2}, replacing ReLU with deterministic LWTA yields negligible differences. On the other hand, stochastic LWTA units yield a clear improvement in all cases. This improvement becomes even more important in the case of our approach, where we sample from stochastic weights.

\begin{table}[h!]
\captionsetup{justification=centering}
\caption{Ablation study ($\%$ classification accuracy)}
\label{table2}
\vskip 0.1in
\centering
\resizebox{\linewidth}{!}{
\begin{tabular}{lccccc}  
\toprule
& & \multicolumn{2}{c}{\textbf{\makecell{Omniglot 20-way}}} &  \multicolumn{2}{c}{\textbf{\makecell{Mini-Imagenet 5-way}}}\\
\midrule
\textbf{Algorithm} & \textbf{Network type} & 1-shot & 5-shot & 1-shot & 5-shot\\
\midrule
MAML (local) & \makecell{deterministic LWTA  \\ stochastic LWTA} & \makecell{95.52  \\ 95.91} & \makecell{98.15  \\ 98.78} & \makecell{48.88  \\ 49.61} & \makecell{63.15  \\ 64.03}   \\
\midrule
FOMAML (local) & \makecell{deterministic LWTA \\ stochastic LWTA}  & \makecell{95.01  \\ 95.80} & \makecell{98.18  \\ 98.41} & \makecell{48.11  \\ 49.24} & \makecell{63.54  \\ 64.54}   \\
\midrule
ABML (local) & \makecell{deterministic LWTA \\ stochastic LWTA}  &  \makecell{90.30  \\ 91.21} & \makecell{93.64  \\ 93.91} & \makecell{44.31  \\ 45.11} & \makecell{52.27  \\ 53.31}   \\
\midrule
BMAML (local) & \makecell{deterministic LWTA \\ stochastic LWTA}  &  \makecell{ 96.96  \\ 97.11 } & \makecell{ 98.21  \\ 98.30 } & \makecell{ 53.12  \\ 53.50} & \makecell{ 64.84\\ 65.31 }   \\
\midrule
PLATIPUS (local) & \makecell{deterministic LWTA \\ stochastic LWTA}  &  \makecell{94.48  \\ 95.13} & \makecell{98.31  \\ 98.56} & \makecell{49.99  \\ 51.06} & \makecell{63.21  \\ 64.18}   \\
\midrule
\textbf{StochLWTA-ML}       & \makecell{deterministic LWTA  \\ stochastic LWTA} & \makecell{96.95 \\ \textbf{97.79}}  & \makecell{98.63 \\ \textbf{98.97}} & \makecell{53.12 \\ \textbf{54.11}}  & \makecell{64.93 \\ \textbf{66.70}}   \\
\bottomrule
\end{tabular}
}
\vskip -0.1in
\end{table}

\subsubsection{Is there a computational time trade-off for the increased accuracy?}
\label{trade_off}
It is also important to investigate whether our approach represents a trade-off between accuracy and computational time compared to our competitors. To facilitate this investigation, in Table \ref{times} we provide training iteration wall-clock times for our approach and the existing locally reproduced state-of-the-art, as well as the total number of iterations each model needs to achieve the reported performance of Table \ref{table1}. It appears that our methodology takes 77$\%$ \emph{less} training time than the less efficient algorithms ABML, BMAML, PLATIPUS, and is comparable to other approaches. This happens because our approach yields the reported state-of-the-art performance by employing a network architecture (that is, number of LWTA layers, as well as number of blocks and block size on each layer) that result in a total number of trainable parameters that is \emph{one order of magnitude less} on average than the best performing baseline methods. This can be seen in the last three columns of Table \ref{times} (dubbed $D_A$, $D_B$ and $D_C$ for Omniglot, Mini-Imagenet and CIFAR-100 respectively). In addition, training for our approach converges fast.

The situation changes when it comes to prediction: our approach imposes a slight computational time overhead compared to MAML, FOMAML and Reptile, but still \emph{much less} than the time-consuming PLATIPUS. BMAML and ABML. This is a rather negligible increase when we are dealing with a low number of drawn samples, $B$. More information on the effect of sample size in our approach’s performance are provided in the Supplementary.

\begin{table*}[h!]
\captionsetup{justification=centering}
\caption{Performance comparison: average wall-clock time (in msecs), training iterations for each locally reproduced method and number of baselines' trainable parameters over the considered datasets of Table \ref{table1}}
\label{times}
\vskip 0.1in
\centering
\resizebox{\linewidth}{!}{
\begin{tabular}{lcccccc}  
\toprule
\textbf{Algorithm}       &  Training & Prediction & \makecell{Training iterations} & $D_A$ parameters & $D_B$ parameters  & $D_C$ parameters\\
\midrule
PLATIPUS (local) &  1603.39 &  602.77  & 333600 & 560025 & 615395  & 580440 \\
BMAML (local) & 1450.31  & 514.43     & 301800 &560025 &  615395  & 580440 \\
ABML (local) & 678.48  &  265.78      & 138000 &224010 & 246158   & 232176 \\
MAML (local)    & 288.25 & 103.28      & 60000 & 112005 &  123079  & 116088 \\
FOMAML (local)  & 284.49 & 102.34      & 60000 & 112005 &  123079  & 116088 \\
Reptile (local) & 284.30 & \textbf{102.27}     & 60000 & 113221 &  124613  & 117463 \\
\midrule
\textbf{StochLWTA-ML} & \textbf{282.90} & 113.44 & 60000 & \textbf{54549}  & \textbf{60112} & \textbf{56745}\\
\bottomrule
\end{tabular}
}
\vskip -0.1in
\end{table*}

\begin{figure*}[h!]
    \centering
    \subfloat[\centering]{{\includegraphics[width=0.40\linewidth]{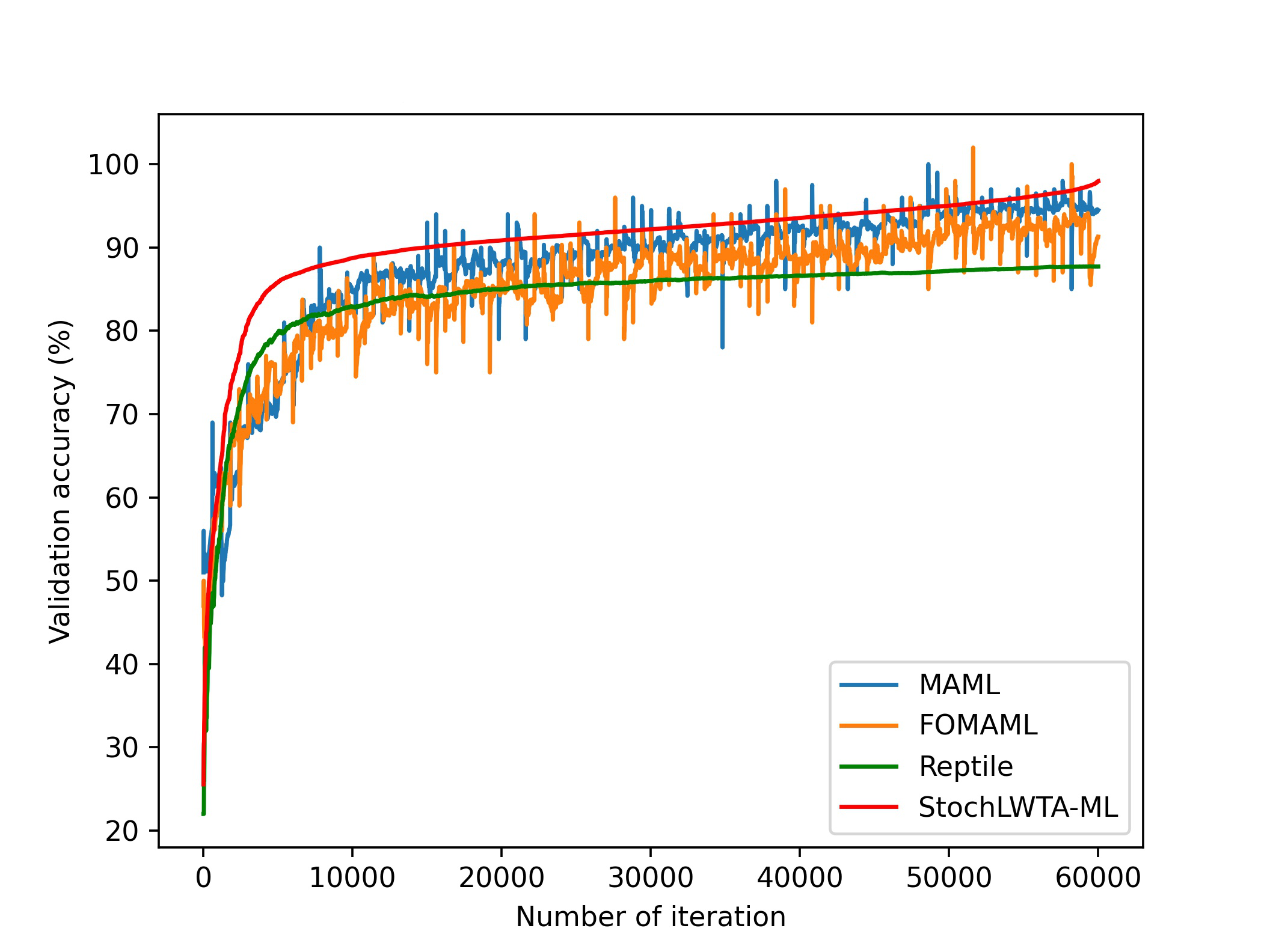} }}
    \qquad
    \subfloat[\centering]{{\includegraphics[width=0.40\linewidth]{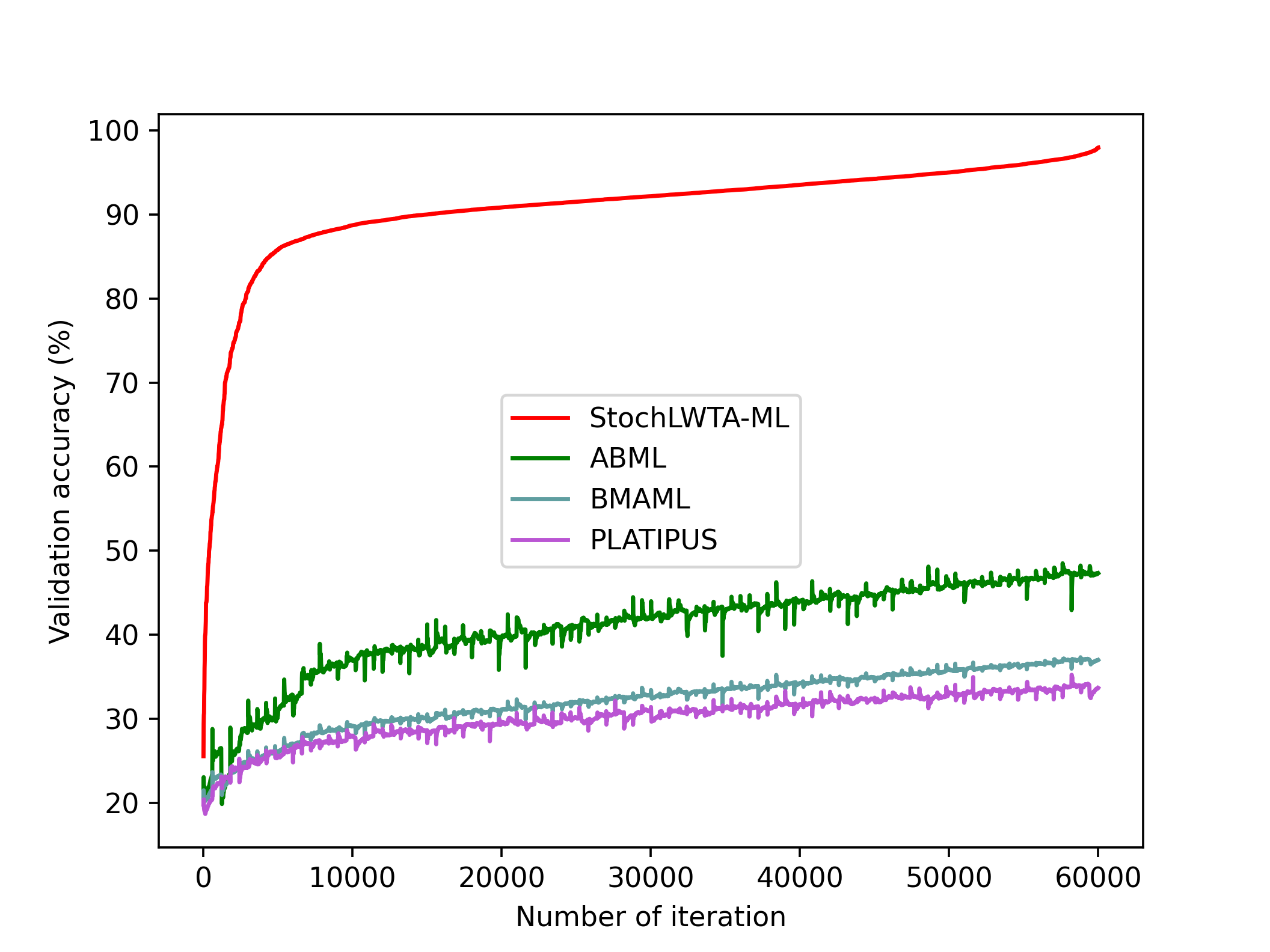} }}
    \caption{ML algorithms' training convergence comparison}
    \label{fig:converge}
    \vskip -0.1in
\end{figure*}

Finally, we provide an example of how training for our approach converges, and how this compares to the alternatives. We illustrate our outcomes on the Omniglot 20-way 1-shot benchmark; similar outcomes have been observed in the rest of the considered datasets. Fig. \ref{fig:converge}(a) compares StochLWTA-ML with prior traditional ML methods: MAML, FOMAML and Reptile. It becomes apparent that our approach converges equally fast to these competitors. Further, Fig. \ref{fig:converge}(b) compares StochLWTA-ML with the probabilistic ML models ABML, BMAML, PLATIPUS. Since these methods are quite time-consuming and less efficient regarding to memory consumption, StochLWTA-ML gives rise to an easier time training MAML based probabilistic model.

\subsubsection{Effect of block size $J$}
As it is presented in Table \ref{units}, increasing the number of competing units per block to $J=4$ or $J=8$ does not notably improve the results of our approach. On the contrary, it increases the number of trained parameters, thus leading to higher network computational complexity. This corroborates our initial choice of using $J=2$ competing units per block in our approach.
 
\begin{table}[h!]
\captionsetup{justification=centering}
\caption{Effect of block size $J$ in StochLWTA-ML's classification ($\%$) accuracy}
\label{units}
\vskip 0.1in
\centering
\resizebox{\linewidth}{!}{
\begin{tabular}{ccccccc}  
\toprule
&  \multicolumn{2}{c}{\textbf{Omniglot 20-way}} &  \multicolumn{2}{c}{\textbf{Mini-Imagenet 5-way}} & \multicolumn{2}{c}{\textbf{CIFAR-100 5-way}}\\
\midrule
\textbf{Number of units}       &  1-shot     &  5-shot    &  1-shot        &  5-shot  &  1-shot        &  5-shot   \\
\midrule
$J=2$                   &  97.79  &  98.97 & 54.11  & 66.70 & 54.60 & 66.73\\
$J=4$                   &  96.33  &  98.55 & 53.99  & 66.65 & 54.51 & 66.13\\
$J=8$                   &  95.38  &  98.83 & 53.70  & 67.08 & 54.45 & 66.18\\
\bottomrule
\end{tabular}
}
\vskip -0.1in
\end{table}

\subsubsection{How does the task batch size affect StochLWTA-ML's performance?}
For demonstration purposes, in Fig. \ref{fig:appendix_ablation}(a) and \ref{fig:appendix_ablation}(b) we illustrate the performance of our model on the Mini-Imagenet 5-way 1-shot setting with different task batch sizes. As we see, our model performs optimally with task batch size of 50 in terms of both training time and predictive accuracy. We have obtained similar results for all other considered datasets.

\subsubsection{How does the sample size $B$ at prediction time affect StochLWTA-ML's accuracy?}
\begin{figure*}[t]
  \vskip 0.1in
  \centering
  \subfloat[\centering]{\includegraphics[width=0.33\linewidth]{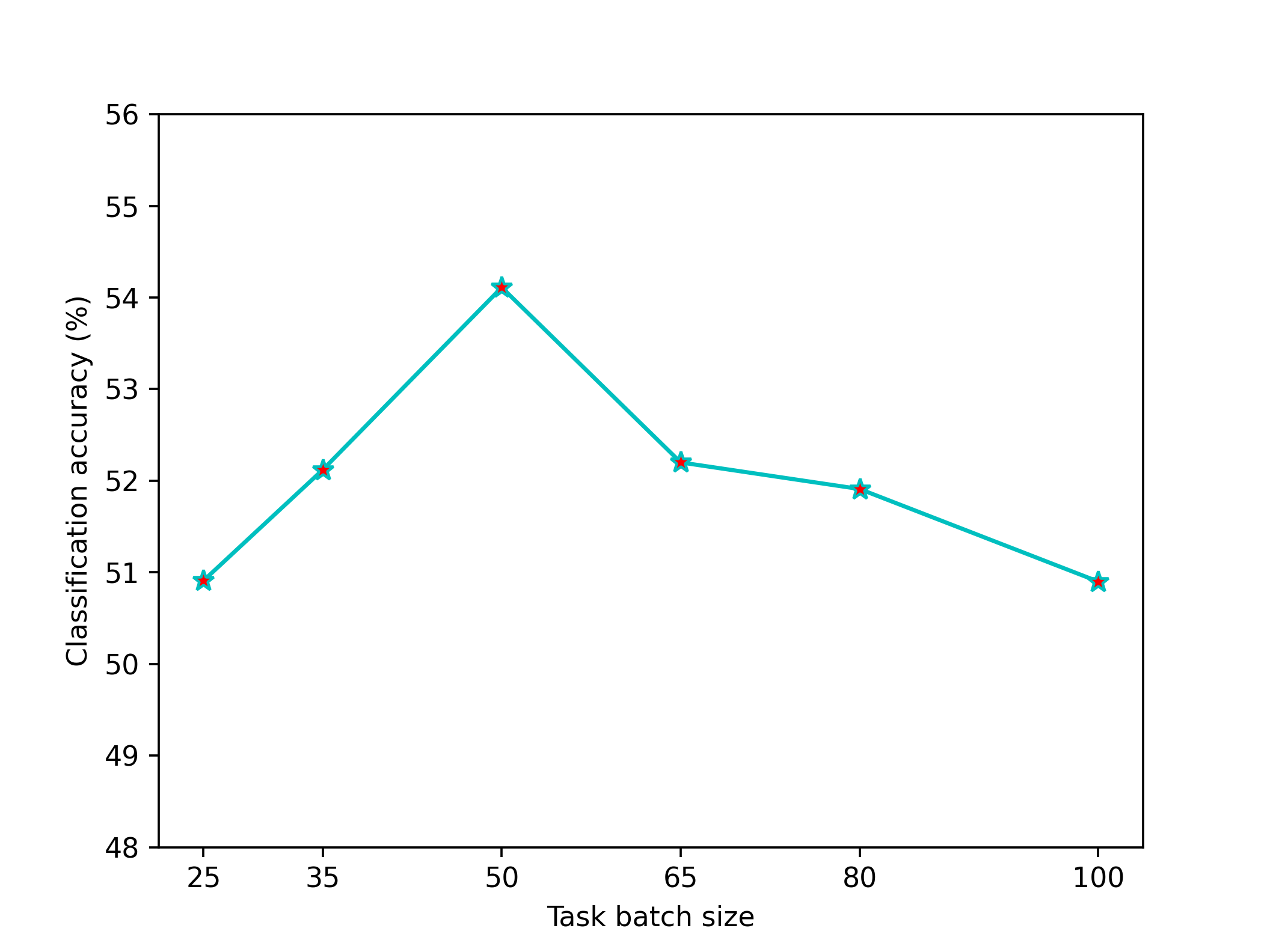}}
  \subfloat[\centering]{\includegraphics[width=0.33\linewidth]{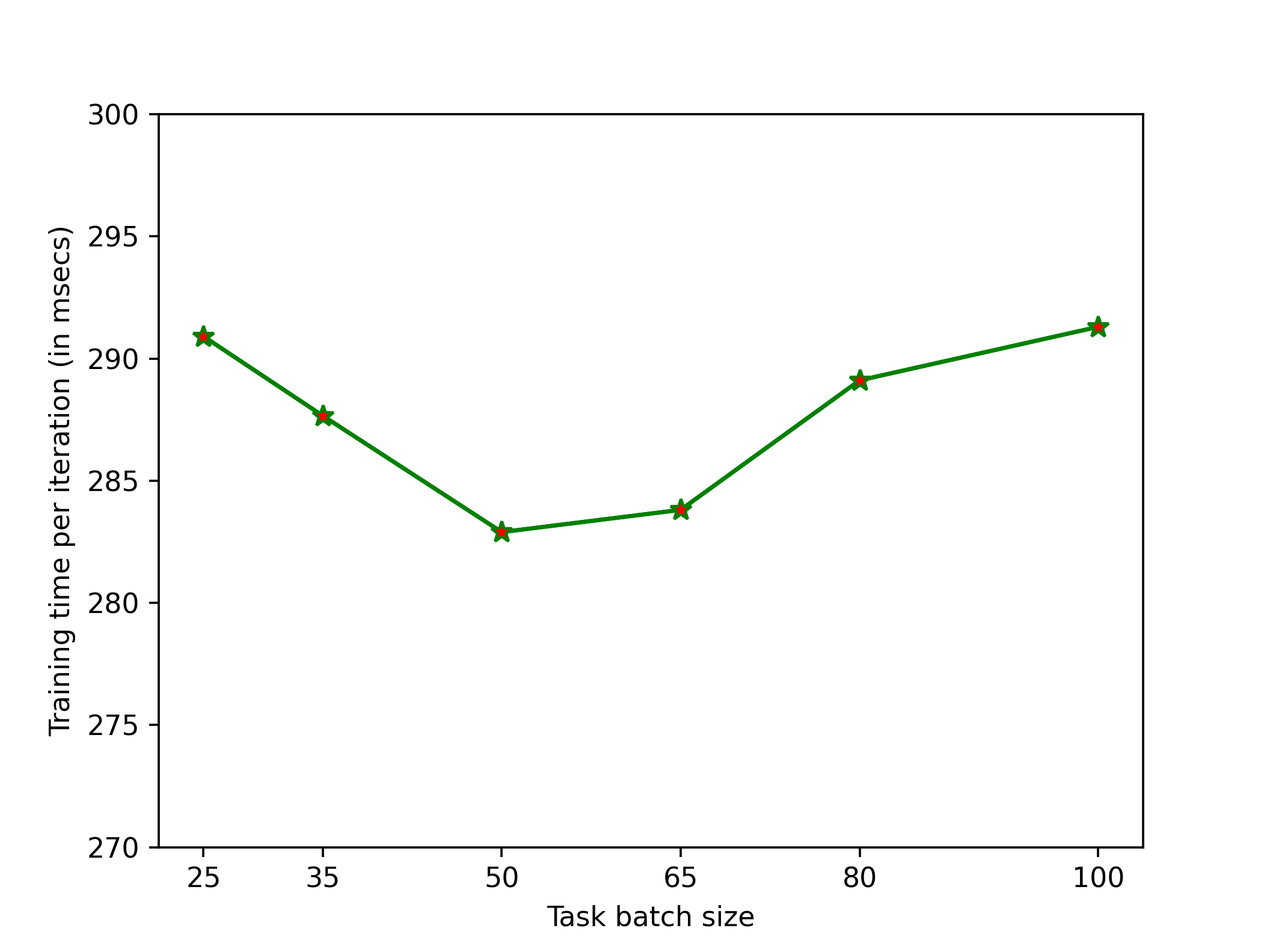}}
  \subfloat[\centering]{\includegraphics[width=0.33\linewidth]{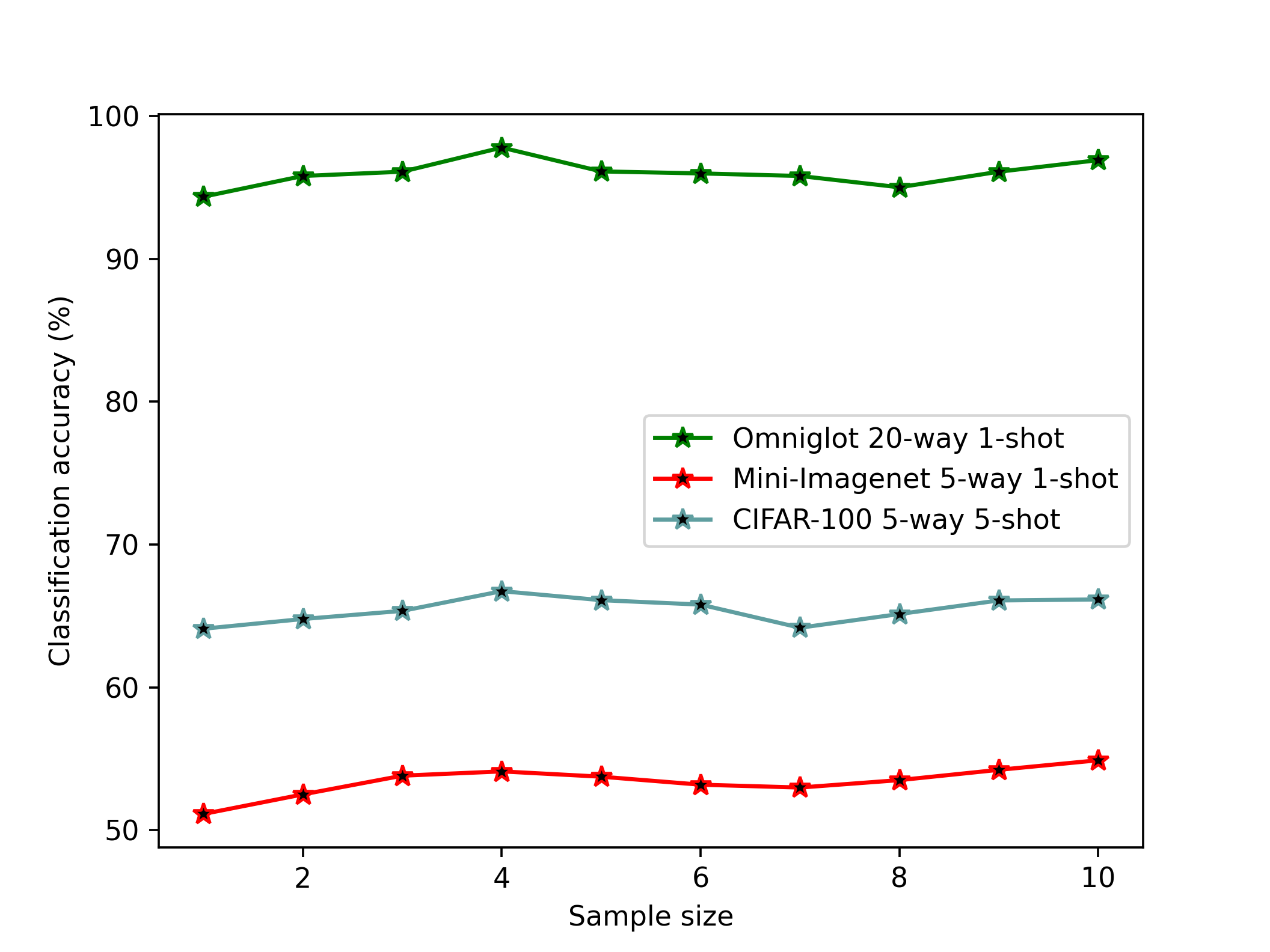}}
  \caption{(a) The effect of task batch size in StochLWTA's predictive accuracy, (b) the effect of task batch size in StochLWTA's training time per iteration (in msecs), and (c) the effect of sample size $B$ in StochLWTA-ML's classification ($\%$) accuracy}
  \label{fig:appendix_ablation}
\end{figure*}

In Fig. \ref{fig:appendix_ablation}(c), we illustrate how sample size $B$ affects predictive accuracy on the Omniglot 20-way 1-shot, Mini-Imagenet 5-way 1-shot and CIFAR-100 5-way 5-shot settings. As we observe, an increase in sample size, $B$, does not always yield an accuracy increase. It seems that selecting $B=4$ allows for the best predictive accuracy/computational complexity trade-off. We have obtained similar findings for the remainder of the considered experimental settings as well. Thus, we finally choose $B=4$ throughout our experiments.

\subsubsection{How important are the stochastic weights?}
We perform an ablation study on Omniglot 20-way; the goal is to discern how much of an extra improvement Gaussian weights offer. Our results are shown in Table \ref{table_point_estimates}. Apparently, stochastic LWTA units offer the greatest fraction of the accuracy gains, but the Gaussian weights are still indispensable.

\begin{table}[h!]
\captionsetup{justification=centering}
\caption{Omniglot 20-way ablation study: Gausssian vs deterministic weights (point estimates).}
\label{table_point_estimates}
\vskip 0.1in
\centering
\begin{tabular}{lcc}  
\toprule
\textbf{Network type} &  \textbf{1-shot} & \textbf{5-shot}\\
\midrule
ReLU (point estimates)            &  95.63 &  96.17 \\
ReLU (Gaussians)                  &  95.80 &  96.48 \\
\midrule
stochastic LWTA (point estimates) & 97.68  &  98.85 \\
\textbf{stochastic LWTA (Gaussians)}   & \textbf{97.79}  &  \textbf{98.97} \\
\bottomrule
\end{tabular}
\vskip -0.1in
\end{table}

\subsection{Regression}
\label{regression}

In this section, we compare our approach StochLWTA-ML with the locally reproduced baselines of Section \ref{classification}, on two sinusoidal function regression problems. In the former case, we apply the default setting used in \citet{Finn}; in the latter, we use a more challenging setting as proposed in \citet{Yoon}, containing more uncertainty than the setting used in \citet{Finn}. Specifically, the tasks distribution $P(T)$ is defined by a sinusoidal function $y=A\sin(\omega x+b) +\epsilon$, with amplitude $A$, frequency $\omega$, phase $b$ and observation noise $\epsilon$. The parameters of each task are sampled from Uniform distributions $A\in[0.1,5.0]$, $b\in[0.0,2\pi]$, $\omega\in[0.5,2.0]$, and observation noise $\epsilon$ from Normal distribution $N(0,(0.01A)^2)$. For each task, 10 input instances $x$ are sampled from $[-5.0,5.0]$. The network architecture employed for the baseline experiments consists of 2 hidden layers of size 64 with tanh activation. In our StochLWTA-ML case, we replace each hidden layer with a proposed stochastic LWTA one. Both architectures end up with a Softmax layer.

\begin{figure*}[h!]
    \centering
    \subfloat[\centering]{{\includegraphics[width=0.33\linewidth]{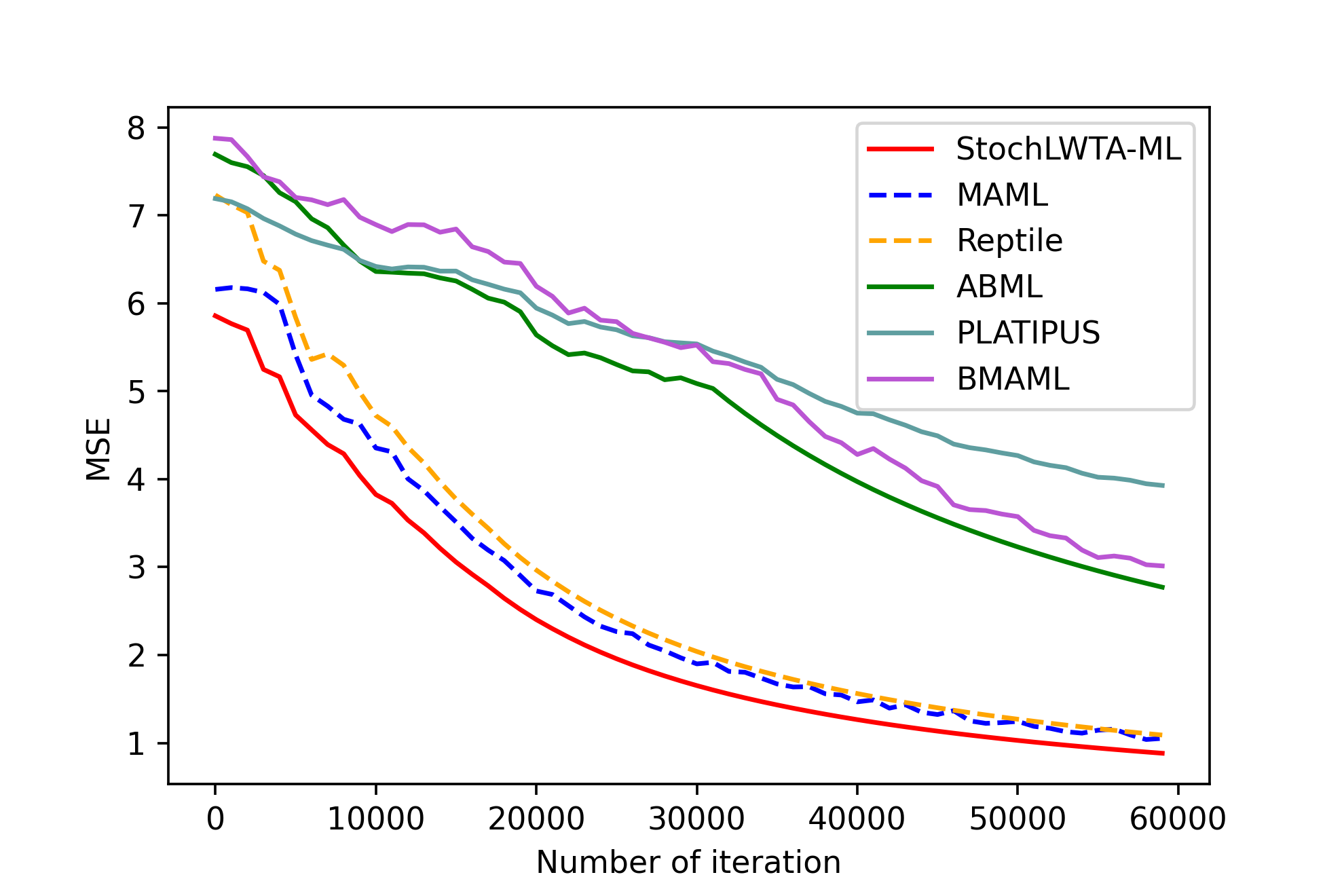} }}
    \subfloat[\centering]{{\includegraphics[width=0.33\linewidth]{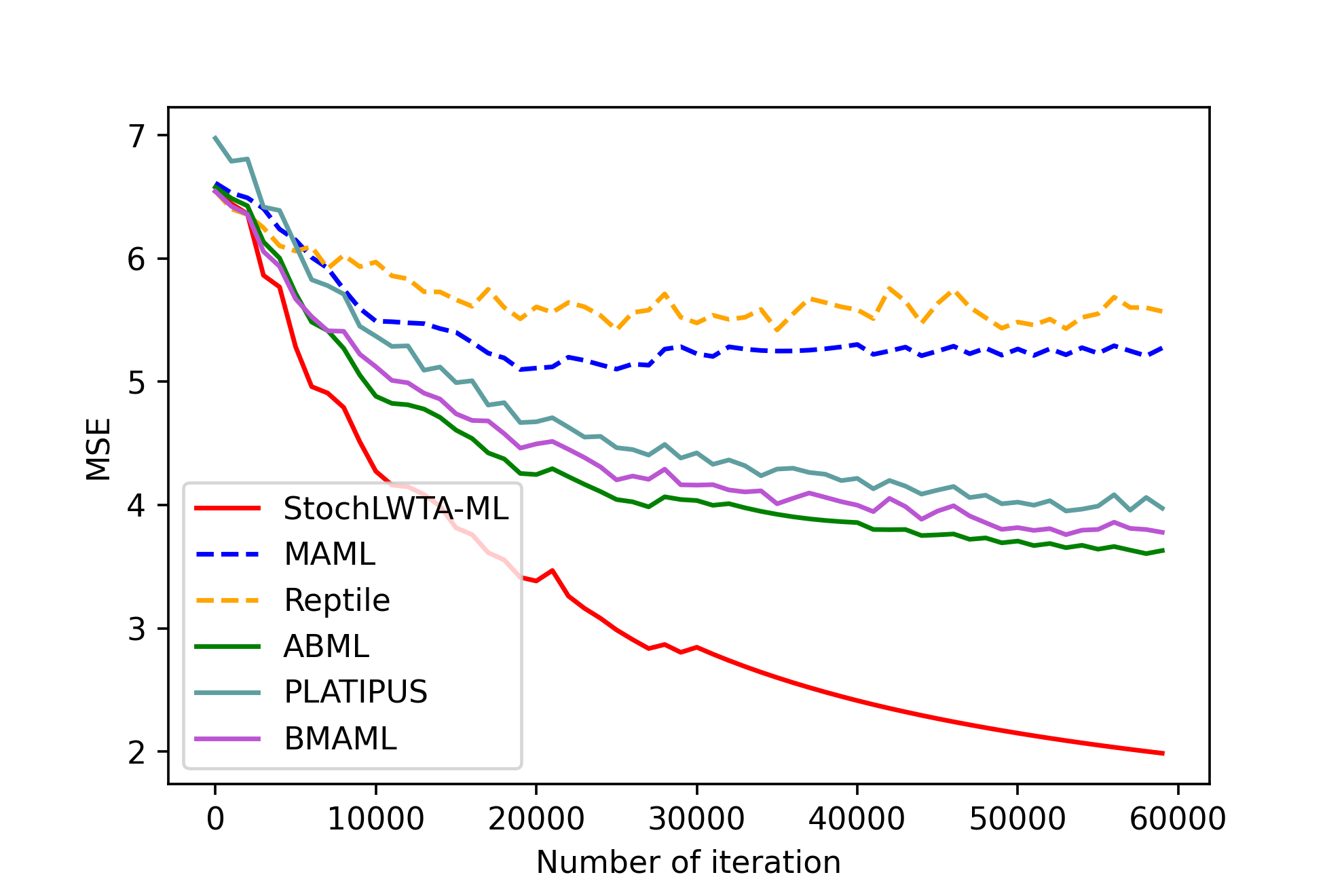} }}
    \subfloat[\centering]{{\includegraphics[width=0.33\linewidth]{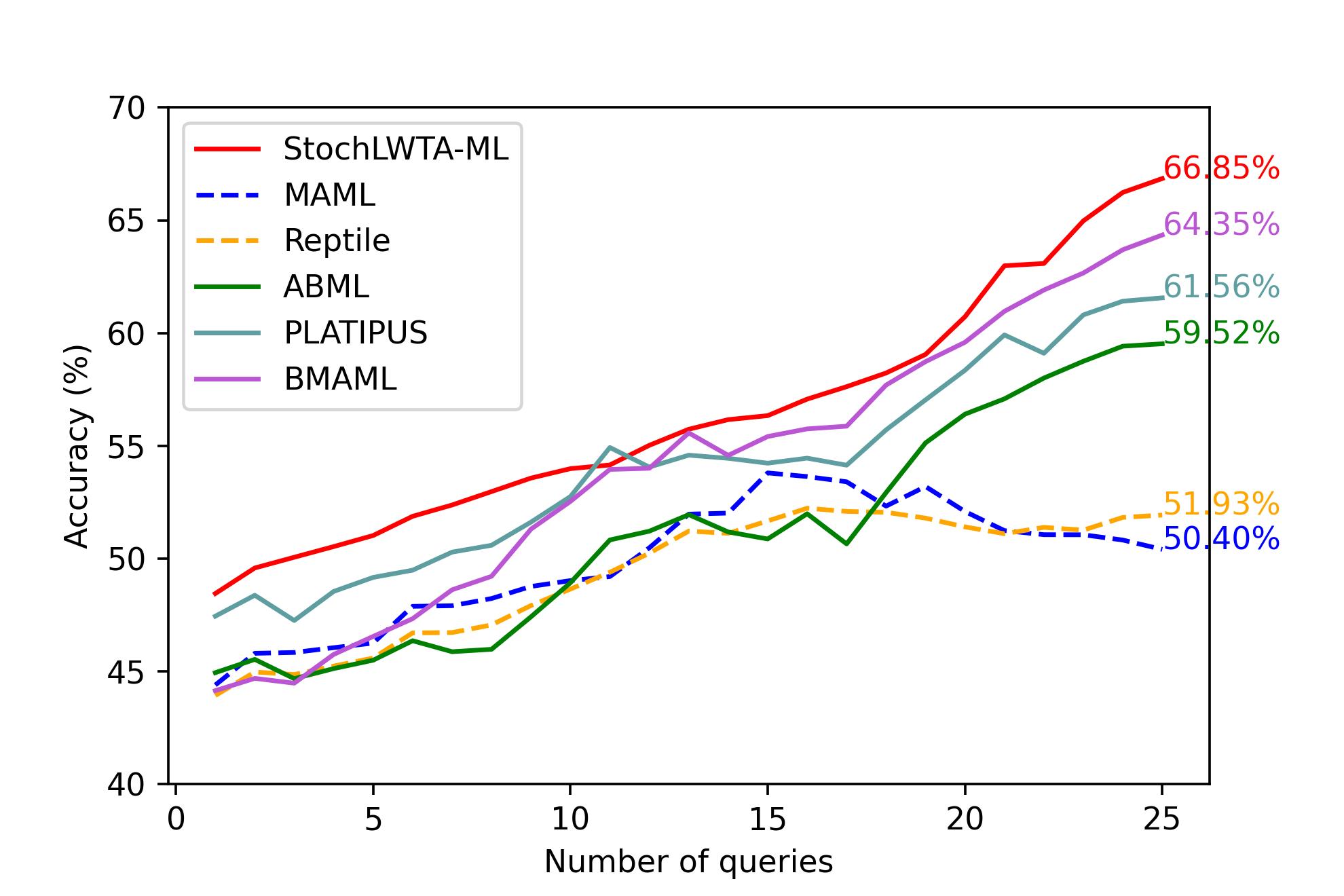} }}
    \caption{Sinusoidal regression results: (a) MSE of default setting after 60000 training iterations, (b) MSE of challenging setting after 60000 training iterations, and (c) active learning setting }
    \label{fig:sinewave}
    \vskip -0.1in
\end{figure*}

In Fig. \ref{fig:sinewave}, we illustrate the Mean Squared Error (MSE) performance on test tasks for both settings, after training all the models for 60000 iterations. Specifically, in Fig. \ref{fig:sinewave}(a) we observe that StochLWTA-ML, in the default setting, adapts equally fast as MAML and Reptile; thus, its convergence speed is much higher than the other time-consuming probabilistic methods ABML, BMAML and PLATIPUS. In the challenging setting of Fig. \ref{fig:sinewave}(b), the Bayesian methods StochLWTA-ML, ABML, BMAML and PLATIPUS can still perform well in a high uncertainty setting while non-Bayesian models MAML and Reptile fail to converge; this outcome is achieved due to the ability of Bayesian methods to reduce overfitting and generalize better. It is clear that our approach yields the most time-efficient method among the baselines. 

\subsubsection{Is there a computational time trade-off for the reduced MSE?}
In Table \ref{sinewave_times}, we report the numbers of trainable parameters as an average over the default and challenging settings of the regression experiments. In a similar vein with the classification outcomes of Section \ref{trade_off}, it appears that our methodology takes 77$\%$ \emph{less} training time than the less efficient algorithms ABML, BMAML, PLATIPUS, and is comparable to other approaches. This demonstrates the effect of parameters' number reduction to the training process of a MAML based probabilistic model.

\begin{table}[h!]
\captionsetup{justification=centering}
\caption{Performance comparison: average wall-clock time (in msecs), and average number of baselines' trainable parameters over the two settings of regression experiments}
\label{sinewave_times}
\vskip 0.1in
\centering
\resizebox{\linewidth}{!}{
\begin{tabular}{lccc}  
\toprule
\textbf{Algorithm}       &  Training & Prediction & Parameters\\
\midrule
PLATIPUS (local) & 359.23  & 135.74  & 21541 \\
BMAML (local) & 323.76 &   115.90   & 21541\\
ABML (local) & 150.96 & 59.75 &   8622\\
MAML (local)    & 65.54  & 23.91 & 4315 \\
FOMAML (local)  & 64.80 &   \textbf{23.02}    &   4315 \\
Reptile (local) & 64.63  & 23.30     &   4353  \\
\midrule
\textbf{StochLWTA-ML} & \textbf{63.11} & 25.45 & \textbf{2092}\\
\bottomrule
\end{tabular}
}
\vskip -0.1in
\end{table}

\subsection{Active learning with regression}
\label{active_learning}

To further evaluate the effectiveness of the proposed stochastic LWTA network, we now consider an active learning experiment, in the challenging setting described in Section \ref{regression}. Specifically, we provide the models with 5 input observations $x$ sampled from $[-5.0,5.0]$. Then, each model queries to label 5 extra instances. The Bayesian methods StochLWTA-ML, ABML, BMAML and PLATIPUS choose an item $x^*$ that has the maximum variance across the sampled regressors. However, the non-Bayesian models MAML and Reptile, choose points randomly, since they do not entail a utility that handles uncertainty; as we demonstrate in Fig. \ref{fig:sinewave}(c), they fail to converge due to their inability to adapt to a more ambiguous training environment. Considering the Bayesian methods, our approach achieves the better predictive performance compared to the other baselines. This reaffirms the prevalence of our proposed network to another experimental scenario.

\section{Conclusion}
\label{conclusion}

In this paper, we proposed a sparse and stochastic network paradigm for ML, with novel network design principles compared to currently used model-agnostic ML models. We introduced \textit{stochastic LWTA activations} in the context of a \textit{variational Bayesian treatment} that gave rise to a doubly-stochastic ML framework, bearing the promise of stronger generalization capacity. We evaluated our approach using standard image classification benchmarks in the field, and showed that it outperformed the state-of-the-art in terms of both predictive accuracy, error rate and computational costs. The results have provided strong empirical evidence supporting our claims. In the future, we plan to study the effect of StochLWTA-ML in areas related to ML, such as Continual Learning \citep{Javed} and Reinforcement Learning \citep{Henry}.

\section*{Acknowledgements}
	
This work has received funding from the European Union's Horizon 2020 research and innovation program under grant agreement No 872139, project aiD.

\bibliography{bibliography}

\begin{thebibliography}{41}
\providecommand{\natexlab}[1]{#1}
\providecommand{\url}[1]{\texttt{#1}}
\expandafter\ifx\csname urlstyle\endcsname\relax
  \providecommand{\doi}[1]{doi: #1}\else
  \providecommand{\doi}{doi: \begingroup \urlstyle{rm}\Url}\fi

\bibitem[Abadi et~al.(2016)Abadi, Agarwal, Barham, Brevdo, Chen, Citro,
  Corrado, Davis, Dean, and Devin]{Abadi}
Abadi, M., Agarwal, A., Barham, P., Brevdo, E., Chen, Z., Citro, C., Corrado,
  G.~S., Davis, A., Dean, J., and Devin, M.
\newblock Tensorflow: Large-scale machine learning on heterogeneous distributed
  systems.
\newblock pp.\  265--283. In \textit{Proceedings of the 12th USENIX Conference
  on Operating Systems Design and Implementation}, 2016.

\bibitem[Andrychowicz et~al.(2016)Andrychowicz, Denil, Gomez, Hoffman, Pfau,
  Schaul, Shillingford, and de~Freitas]{Marcin}
Andrychowicz, M., Denil, M., Gomez, S., Hoffman, M.~W., Pfau, D., Schaul, T.,
  Shillingford, B., and de~Freitas, N.
\newblock Learning to learn by gradient descent by gradient descent.
\newblock In \textit{Neural Information Processing Systems}, 2016.

\bibitem[Baik et~al.(2020)Baik, Choi, Choi, Kim, and Lee]{Sungyong}
Baik, S., Choi, M., Choi, J., Kim, H., and Lee, K.~M.
\newblock Meta-learning with adaptive hyperparameters.
\newblock In \textit{Neural Information Processing Systems}, 2020.

\bibitem[Black et~al.(2006)Black, McCormick, James, and Pedderd]{Black}
Black, P., McCormick, R., James, M., and Pedderd, D.
\newblock Learning how to learn and assessment for learning: a theoretical
  inquiry.
\newblock \emph{Research Papers in Education}, 21:\penalty0 119--132, 2006.

\bibitem[Buysse et~al.(2019)Buysse, Mets, and Latré]{Matthias}
Buysse, H.~M., Mets, K., and Latré, S.
\newblock Fast task-adaptation for tasks labeled using natural language in
  reinforcement learning.
\newblock 2019.
\newblock URL \url{https://arxiv.org/abs/1910.04040}.

\bibitem[Castro et~al.(2008)Castro, Kalish, Nowak, Qian, Rogers, and
  Zhu]{Castro}
Castro, R., Kalish, C., Nowak, R., Qian, R., Rogers, T., and Zhu, J.
\newblock Human active learning.
\newblock In \textit{Neural Information Processing Systems}, 2008.

\bibitem[Chen et~al.(2020)Chen, Friesen, Behbahani, Doucet, Budden, Hoffman,
  and de~Freitas]{Chen}
Chen, Y., Friesen, A.~L., Behbahani, F., Doucet, A., Budden, D., Hoffman, M.,
  and de~Freitas, N.
\newblock Modular meta-learning with shrinkage.
\newblock In \textit{Neural Information Processing Systems}, 2020.

\bibitem[Edwards \& Storkey(2016)Edwards and Storkey]{Harrison}
Edwards, H. and Storkey, A.
\newblock Towards a neural statistician.
\newblock 2016.
\newblock URL \url{https://arxiv.org/abs/1606.02185}.

\bibitem[Finn et~al.(2017)Finn, Abbeel, and Levine]{Finn}
Finn, C., Abbeel, P., and Levine, S.
\newblock Model-agnostic meta-learning for fast adaptation of deep networks.
\newblock In \textit{International Conference on Machine Learning}, 2017.

\bibitem[Finn et~al.(2018)Finn, Xu, and Levine]{Finn_Xu}
Finn, C., Xu, K., and Levine, S.
\newblock Probabilistic model-agnostic meta-learning.
\newblock In \textit{Neural Information Processing Systems}, 2018.

\bibitem[Finn et~al.(2019)Finn, Rajeswaran, Kakade, and Levine]{Finn2}
Finn, C., Rajeswaran, A., Kakade, S., and Levine, S.
\newblock Online meta-learning.
\newblock In \textit{International Conference on Machine Learning}, 2019.

\bibitem[Glorot \& Bengio(2010)Glorot and Bengio]{glorot}
Glorot, X. and Bengio, Y.
\newblock Understanding the difficulty of training deep feedforward neural
  networks.
\newblock \emph{Journal of Machine Learning Research 9}, pp.\  249--256, 2010.

\bibitem[Gordon et~al.(2018)Gordon, Bronskill, Bauer, Nowozin, and
  Turner]{Gordon_Bronskill}
Gordon, J., Bronskill, J., Bauer, M., Nowozin, S., and Turner, R.~E.
\newblock Meta-learning probabilistic inference for prediction.
\newblock In \textit{International Conference on Learning Representations},
  2018.

\bibitem[Grant et~al.(2018)Grant, Finn, Levine, Darrell, and
  Griffiths]{Grant_Finn}
Grant, E., Finn, C., Levine, S., Darrell, T., and Griffiths, T.
\newblock Recasting gradient-based meta-learning as hierarchical bayes.
\newblock In \textit{International Conference on Learning Representations},
  2018.

\bibitem[Ioffe \& Szegedy(2015)Ioffe and Szegedy]{Ioffe}
Ioffe, S. and Szegedy, C.
\newblock Batch normalization: Accelerating deep network training by reducing
  internal covariate shift.
\newblock In \textit{International Conference on Machine Learning}, 2015.

\bibitem[Javed \& White(2019)Javed and White]{Javed}
Javed, K. and White, M.
\newblock Meta-learning representations for continual learning.
\newblock pp.\  1818--1828. In \textit{Neural Information Processing Systems},
  2019.

\bibitem[Kessler et~al.(2021)Kessler, Nguyen, Zohren, and Roberts]{Kessler}
Kessler, S., Nguyen, V., Zohren, S., and Roberts, S.
\newblock Hierarchical indian buffet neural networks for bayesian continual
  learning.
\newblock In \textit{UAI}, 2021.

\bibitem[Kingma \& Welling(2014)Kingma and Welling]{Kingma}
Kingma, D.~P. and Welling, M.
\newblock Auto-encoding variational bayes.
\newblock In \textit{International Conference on Learning Representations},
  2014.

\bibitem[Krizhevsky(2009)]{Krizhevsky_cifar}
Krizhevsky, A.
\newblock Learning multiple layers of features from tiny images.
\newblock Technical report, 2009.

\bibitem[Kühl et~al.(2020)Kühl, Goutier, Baier, Wolff, and Martin]{Kuhl}
Kühl, N., Goutier, M., Baier, L., Wolff, C., and Martin, D.
\newblock Human vs. supervised machine learning: Who learns patterns faster?
\newblock 2020.
\newblock URL \url{https://arxiv.org/abs/2012.03661}.

\bibitem[Lai(2019)]{Lai}
Lai, Y.
\newblock A comparison of traditional machine learning and deep learning in
  image recognition.
\newblock \emph{Journal of Physics: Conference Series}, 1314, 2019.

\bibitem[Lake et~al.(2017)Lake, Ullman, Tenenbaum, and Gershman]{Lake}
Lake, B.~M., Ullman, T.~D., Tenenbaum, J.~B., and Gershman, S.~J.
\newblock Building machines that learn and think like people.
\newblock \emph{Behavioral and Brain Sciences}, 40, 2017.

\bibitem[Lee et~al.(2018)Lee, Kim, Yoon, Lee, Yang, and Hwang]{Lee_Kim}
Lee, J., Kim, S., Yoon, J., Lee, H.~B., Yang, E., and Hwang, S.~J.
\newblock Adaptive network sparsification with dependent variational
  beta-bernoulli dropout.
\newblock 2018.
\newblock URL \url{https://arxiv.org/abs/1805.10896}.

\bibitem[Maddison et~al.(2017)Maddison, Mnih, and Teh]{Maddison}
Maddison, C.~J., Mnih, A., and Teh, Y.~W.
\newblock The concrete distribution: A continuous relaxation of discrete random
  variables.
\newblock In \textit{International Conference on Learning Representations},
  2017.

\bibitem[Nalisnick et~al.(2021)Nalisnick, Gordon, and
  Hernández-Lobato]{Nalisnick}
Nalisnick, E., Gordon, J., and Hernández-Lobato, J.~M.
\newblock Predictive complexity priors.
\newblock In \textit{Neural Information Processing Systems}, 2021.

\bibitem[Nichol et~al.(2018)Nichol, Achiam, and Schulman]{Nichol}
Nichol, A., Achiam, J., and Schulman, J.
\newblock On first-order meta-learning algorithms.
\newblock 2018.
\newblock URL \url{https://arxiv.org/abs/1803.02999}.

\bibitem[Panousis et~al.(2021)Panousis, Chatzis, Alexos, and
  Theodoridis]{Panousis_Chatzis_Adv}
Panousis, K., Chatzis, S., Alexos, A., and Theodoridis, S.
\newblock Local competition and stochasticity for adversarial robustness in
  deep learning.
\newblock In \textit{International Conference on Artificial Intelligence and
  Statistics}, 2021.

\bibitem[Panousis et~al.(2019)Panousis, Chatzis, and
  Theodoridis]{Panousis_Chatzis}
Panousis, K.~P., Chatzis, S., and Theodoridis, S.
\newblock Nonparametric bayesian deep networks with local competition.
\newblock In \textit{International Conference on Machine Learning}, 2019.

\bibitem[Patacchiola et~al.(2020)Patacchiola, Turner, Crowley, O’Boyle, and
  Storkey]{Patacchiola}
Patacchiola, M., Turner, J., Crowley, E.~J., O’Boyle, M., and Storkey, A.
\newblock Bayesian meta-learning for the few-shot setting via deep kernels.
\newblock In \textit{Neural Information Processing Systems}, 2020.

\bibitem[Ravi \& Beatson(2019)Ravi and Beatson]{Ravi_Beatson}
Ravi, S. and Beatson, A.
\newblock Amortized bayesian meta-learning.
\newblock In \textit{International Conference on Learning Representations},
  2019.

\bibitem[Ravi \& Larochelle(2017)Ravi and Larochelle]{Ravi}
Ravi, S. and Larochelle, H.
\newblock Optimization as a model for few-shot learning.
\newblock In \textit{International Conference on Learning Representations},
  2017.

\bibitem[Robbins(2007)]{Robbins}
Robbins, H.
\newblock A stochastic approximation method.
\newblock \emph{Annals of Mathematical Statistics}, 2007.

\bibitem[Santoro et~al.(2016)Santoro, Bartunov, Botvinick, Wierstra, and
  Lillicrap]{Santoro}
Santoro, A., Bartunov, S., Botvinick, M., Wierstra, D., and Lillicrap, T.
\newblock Meta-learning with memory-augmented neural networks.
\newblock In \textit{International Conference on Machine Learning}, 2016.

\bibitem[Sculley et~al.(2015)Sculley, Holt, Golovin, Davydov, and
  Phillips]{Sculley}
Sculley, D., Holt, G., Golovin, D., Davydov, E., and Phillips, T.
\newblock Hidden technical debt in machine learning systems.
\newblock In \textit{Neural Information Processing Systems}, 2015.

\bibitem[Srivastava et~al.(2013)Srivastava, Masci, Kazerounian, Gomez, and
  Schmidhube]{Rupesh}
Srivastava, R.~K., Masci, J., Kazerounian, S., Gomez, F., and Schmidhube, J.
\newblock Compete to compute.
\newblock In \textit{Neural Information Processing Systems}, 2013.

\bibitem[Triantafillou et~al.(2017)Triantafillou, Zemel, and
  Urtasun]{Triantafillou}
Triantafillou, E., Zemel, R., and Urtasun, R.
\newblock Few-shot learning through an information retrieval lens.
\newblock In \textit{Neural Information Processing Systems}, 2017.

\bibitem[Vinyals et~al.(2016)Vinyals, Blundell, Lillicrap, Kavukcuoglu, and
  Wierstra]{Vinyals}
Vinyals, O., Blundell, C., Lillicrap, T., Kavukcuoglu, K., and Wierstra, D.
\newblock Matching networks for one shot learning.
\newblock In \textit{Neural Information Processing Systems}, 2016.

\bibitem[Voskou et~al.(2021)Voskou, Panousis, Kosmopoulos, Metaxas, and
  Chatzis]{Voskou_Chatzis}
Voskou, A., Panousis, K., Kosmopoulos, D., Metaxas, D., and Chatzis, S.
\newblock Stochastic transformer networks with linear competing units:
  Application to end-to-end sl translation.
\newblock In \textit{International Conference on Computer Vision}, 2021.

\bibitem[Yoon et~al.(2018)Yoon, Kim, Dia, Kim, Bengio, and Ahn]{Yoon}
Yoon, J., Kim, T., Dia, O., Kim, O., Bengio, Y., and Ahn, S.
\newblock Bayesian model-agnostic meta-learning.
\newblock In \textit{Neural Information Processing Systems}, 2018.

\bibitem[Zhu et~al.(2020)Zhu, Yu, Gupta, Shah, Hartikainen, Singh, Kumar, and
  Levine]{Henry}
Zhu, H., Yu, J., Gupta, A., Shah, D., Hartikainen, K., Singh, A., Kumar, V.,
  and Levine, S.
\newblock The ingredients of real-world robotic reinforcement learning.
\newblock In \textit{International Conference on Learning Representations},
  2020.

\bibitem[Zou \& Lu(2020)Zou and Lu]{Zou_Lu}
Zou, Y. and Lu, X.
\newblock Gradient-em bayesian meta-learning.
\newblock In \textit{Neural Information Processing Systems}, 2020.

\end{thebibliography}
\bibliographystyle{icml2022}

\newpage
\appendix

\section{Further details on the used datasets}

Omniglot is a dataset of 1623 characters from different alphabets, containing 20 examples per character scaled down to 28x28 grayscale pixels. The ratio between training and testings sets is 3:2, so after shuffling the character classes we randomly choose the first 974 classes for training and the remaining are left for testing. As for the Mini-Imagenet dataset, it has color images of size 84x84 and contains 100 classes with 600 examples from the ImageNet dataset. We randomly choose 45000 examples for the training phase and the rest constitute the testing population. The CIFAR-100 dataset consists of color images of size 32x32 and contains 100 classes with 600 images per class. We randomly choose 500 images per class for training and the rest 100 images per class constitute the testing population.

\section{Few-Shot Classification Network Architectures}

For the local replicates of prior ML algorithms in the experiments of our work, we follow the same architecture for the deep neural network as the one used by \citet{Vinyals}. For Omniglot, the network is composed of 4 convolutional layers with 64 filters, 3 x 3 convolutions and 2 x 2 strides, followed by a Batch Normalization layer \citep{Ioffe} and the final values of each layer are processed by an activation function. For both Mini-Imagenet and CIFAR-100, we use 4 convolutional layers with 32 filters to reduce overfitting like \citet{Ravi}, 3 x 3 convolutions followed by  Batch Normalization layer and 2 × 2 max-pooling layer with the values of each layer finally passed again through an activation block. The activation function used for the baseline experiments is: ReLU for Tables \ref{table1}, \ref{times}, \ref{sinewave_times} and \ref{sota_same_params}, and LWTA for Table \ref{table2}.

\section{What parameters do we count for the outcomes of Tables \ref{times}, \ref{sinewave_times} and \ref{sota_same_params}?}
The included parameters of each baseline for the outcomes of Tables \ref{times}, \ref{sinewave_times} and \ref{sota_same_params} are:
\setcounter{table}{0}
\renewcommand{\thetable}{D\arabic{table}}

\begin{itemize}
\item PLATIPUS: $\Theta = \{\bm{\mu}_{\theta}, \bm{\sigma}^2_{\theta}, \bm{v}_{\theta}, \bm{\gamma}_p, \bm{\gamma}_q\}$
\item BMAML: $\Theta=\{\theta^m\}_{m=1}^5$, for using 5 particles
\item ABML: $\theta=\{\bm{\mu}_{\theta},\bm{\sigma}^2_{\theta}\}$
\item MAML: $\theta=\{\bm{\mu}_{\theta}\}$
\item FOMAML: $\theta=\{\bm{\mu}_{\theta}\}$
\item Reptile: $\theta=\{\bm{\mu}_{\theta}\}$
\item StochLWTA-ML: $\theta=\{\bm{\mu}_{\theta},\bm{\sigma}^2_{\theta}\}$
\end{itemize}

\begin{table*}[h!]
\captionsetup{justification=centering}
\caption{Performance comparison: wall-clock time (in msecs), training iterations for each locally reproduced method, classification accuracy and number of baselines' trainable parameters over the Omniglot 20-way 1-shot benchmark}
\label{sota_same_params}
\vskip 0.1in
\centering
\begin{tabular}{lccccc}  
\toprule
\textbf{Algorithm}       &  Training & Prediction & \makecell{Training iterations} & Accuracy $(\%)$ &  Parameters \\
\midrule
PLATIPUS (local) & 490.67 &  221.91  & 107100  & 91.57 & 55817  \\
BMAML (local) & 479.03  &  200.68    & 103560  & 94.11 & 56321  \\
ABML (local) & 402.38  &    170.32   &  87000  & 87.92 & 55312  \\
MAML (local)    & 272.89 &  91.24     & 60000 & 92.10 & 55917  \\
FOMAML (local)  & 269.01 & 91.12      & 60000 & 92.83 & 55917  \\
Reptile (local) & \textbf{268.72} & \textbf{90.83}  & 60000 & 85.60 & 56525  \\
\midrule
\textbf{StochLWTA-ML} & 272.14 & 102.56 & 60000 & \textbf{97.79} & \textbf{54549}\\
\bottomrule
\end{tabular}
\vskip -0.1in
\end{table*}

\section{More Ablations} 
\subsection{How do the state-of-the-art methods perform with a parameter count reduced to be about the same as StochLWTA-ML?}

We repeated Omniglot 20-way 1-shot experiments to evaluate all state-of-the-art, using a deep network of the same number of parameters and similar architecture as the proposed StochLWTA-ML model. To this end, we simply replaced the stochastic LWTA layers with dense ReLU layers of the same size, and dropped the Gaussians from the weights. This yielded between 2.2$\%$ and 3.5$\%$ reduction in classification accuracy, as we show in Table \ref{sota_same_params}.

\subsection{How does StochLWTA-ML perform with more parameters?}
\setcounter{table}{0}
\renewcommand{\thetable}{E\arabic{table}}
We repeated a classification experiment on Omniglot 20-way 1-shot by using a stochastic LWTA architecture of 4 convolutional layers. In this context, in each layer competition is performed among the feature maps of a convolutional kernel; this proceeds on a position-wise basis. The so-obtained convolutional deep network yields the same number of parameters as in the state-of-the-art. Our experimental outcomes are conspicuous: our approach lost accuracy (the reported 97.79$\%$ accuracy reduced to 96.50$\%$), while training time increased by four and inference time almost doubled; similar outcomes have been observed in the rest of the considered datasets.

\end{document}